# Symmetry Detection of Occluded Point Cloud Using Deep Learning

Zhelun Wu[a]*, Hongyan Jiang[b], Siyun He[c]

[a] *Shenzhen Institute of Advanced Technology, Chinese Academy of Science, China*
[b] *China Telecom Tech Division Hunan, China*
[c] *Yale University, New Haven, United States*

**Abstract**

Symmetry detection has been a classical problem in computer graphics, many of which using traditional geometric methods. In recent years, however, we have witnessed the arising deep learning changed the landscape of computer graphics. In this paper, we aim to solve the symmetry detection of the occluded point cloud in a deep-learning fashion. To the best of our knowledge, we are the first to utilize deep learning to tackle such a problem. In such a deep learning framework, double supervisions: points on the symmetry plane and normal vectors are employed to help us pinpoint the symmetry plane. We conducted experiments on the YCB-video dataset and demonstrate the efficacy of our method. Our code will be available soon after on GitHub.

## 1. Introduction

In the human visual system, our brains are very skilled at finding and utilizing symmetry of objects as abstraction entails finding intrinsic logic of shape. Such an assumption is legitimate, given the ubiquity of symmetry in human-made merchandise.

Additionally, symmetry offers a critical hint when it comes to object completion in occlusion. Completion of such is conducive to object detection and makes it more robust in cluttered scenarios. Despite all these advantages in mind, the status quo is far from satisfactory. Most works on symmetry are built on top of classical methodology, including curvature, ICP (Iterative Closest Point), and others. None had dealt with deep-learning oriented symmetry detection on occluded point cloud before, save for PRS-Net, which aimed to predict symmetry plane for the mesh-based 3D objects with deep-learning techniques.

Unlike PRS-Net with exact three symmetry predictions, our algorithm can deal with an indefinite amount of symmetries as we base on double supervisions and RANSAC (random sample consensus). In our work, we are trying to locate the symmetry plane by two elements: the points on the plane and the normal vector of the plane. These are the very two elements that determine where a plane is.

Two stages of training are employed in order to get our model well-trained. In the first stage, the primary goal is to let the model learn the whereabouts of those points. The second stage begins when we ascertain the stability of point predictions. In the second stage, the model learns to predict normal vectors concerning those points. After these two stages, we can extrapolate the location of the symmetry plane using RANSAC.

Actually, by introducing RANSAC into the algorithm, we successfully do away with the restriction of fixation on the number of predictions. Of course, we are not saying our algorithm is robust enough. For one, our method relies on the presence of points on symmetry planes. Further works on more robust algorithms may well become available after this paper.

This paper mainly has three contributions:
1. We propose a novel deep-learning symmetry detection framework for 3D models. Our model is **the first** effective model to deal with the occluded point cloud. By using double supervisions, we can detect the

* Corresponding author. Tel.: +86-13618490805.
 *E-mail address:* zhelunwu@hotmail.com

     symmetry as long as points on the symmetry plane are present in the observer's view.
2. Attributed to a two-step framework and RANSAC, our algorithm allows flexible numbers of symmetry to be learned and detected under the framework.
3. A new accuracy measurement for symmetry detection is designed, and we elaborate on how PR-curve can be drawn in our scenario.

## 2. Related Works

### 2.1. Symmetry Detection

Both 2D and 3D geometry have a long history of symmetry detection. Generally, there are four types of symmetry problems, two types on two dimensions: extrinsic and intrinsic, global and partial.

Global and partial symmetry differ on the scale of symmetry; the former requires the symmetry to map the entire object to itself while the latter only requires part of the object. Extrinsic and intrinsic symmetry differ on metrics of the symmetry; the previous measures symmetry on natural Euclidean space and the other measures on different metric spaces. In this paper, we are aiming to solve extrinsic global symmetry detection, especially for reflective symmetry detection.

As far as extrinsic global symmetry concerns, various works have made progress. Some researchers have made adaptions to the original problem. Zabrodsky et al.[1] introduce the concept of approximate symmetry to tackle the problem. Raviv et al.[2] adapt the symmetries for non-rigid shapes and Kazhdan et al.[3] offer to solve n-fold rotational symmetries. Others use mapping to solve the problem. Podolak et al.[4] use PRST (planar reflective symmetry transform) to detect symmetry. Mitra et al.[6] also use transformation to extract constant symmetry features and clustering to detect symmetries. However, most of these works are aimed to tackle 2D detection.

Other works can handle 3D symmetry detections. One of the closest is the work by Aleksandrs Ecins et al.[5] where cluttered objects are segmented by help of recursively detecting symmetries and segmenting. We will compare the effect of this paper with our paper in the experiments. PRS-Net[7] is the first one to deal with 3D data with deep learning methods; however, unlike our paper, it is not designed to adapt to cluttered objects.

### 2.2. Point Cloud Deep Representation

3D cases are quite different from 2D images. The geometry and depth information need to be absorbed in conquering symmetry detection. The complexity of 3D geometry representation probably explains the diversity when it comes to dealing with symmetry detections. PRS-Net[7] deals with mesh-based 3D models. Paper[5] by Ecins uses point clouds; however, it also requires occlusion map[9]. Our paper does not require extra information other than RGB-D images. The framework is heavily dependent on the DenseFusion[8] framework. DenseFusion extracts point cloud information from multiple scales by leveraging both color and geometry information.

# 3. Methods

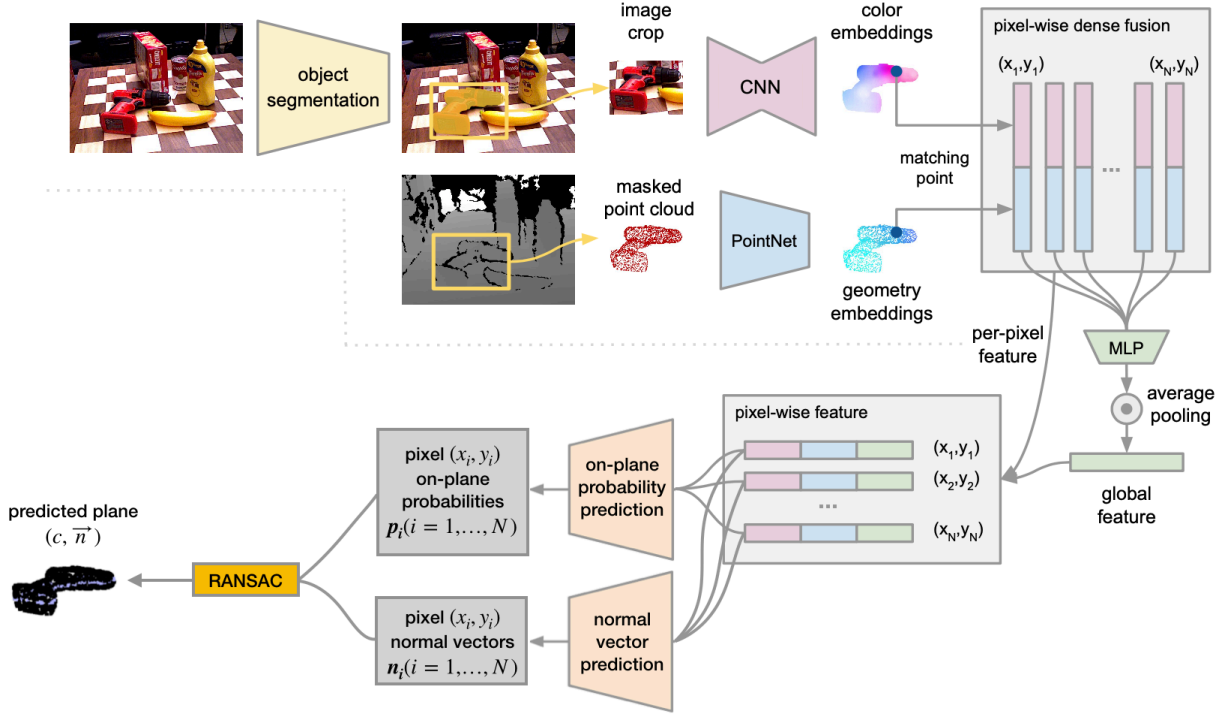

Fig. 1. Symmetry Detection Framework Using Double Supervision

Customarily, when we predict reflectional symmetry, we are anticipating a point on the plane and the normal vector. In this method, we conquer the reflection symmetry detection of occluded point cloud by leveraging two point-wise predictions. First, we learn to predict the on-plane confidence, whether a point is on the symmetry plane. Meanwhile, point-wise normal vectors are predicted for on-plane points. Since the location of each on-plane point combined with its normal vector prediction uniquely decides the symmetry plane, we should fuse all these symmetry plane predictions of on-plane points into our final symmetry predictions. In this case, we employ RANSAC to fuse point-wise predictions.

## 3.1. Feature Extraction

In our framework, as shown in Fig. 1, we adopted the DenseFusion framework as our backbone in geometric and color embedding extraction. In the first stage, we use the YCB ground-truth segmentation label to segment the object out on the photo. The segmented object is then projected onto the 3D space. Apparently, these projections may be occluded, even heavily occluded. These projected points are passed into the network, and CNN (convolutional neural network) is employed to extract color embeddings while resorting to PointNet [10] for geometric embeddings. These pixel-wise features are then fused and concatenated together, leveraged to generate on-plane confidence and point-wise normal vectors.

## 3.2. Training

To successfully train the model, we first let the point learn if it is on the symmetry plane. Those points that are closer than $\epsilon_1$ to the symmetry plane are regarded as on-plane points. Those on-plane points should be labeled as **1**, while off-plane points should be labeled as **0**. We adopt cross-entropy loss here:

$$L_{op} = \sum_{S \in Sym} \left( - \sum_{p \in O(S) \wedge P(p) \leq P_{th}} \log P(p) - \sum_{p \notin O(S) \wedge P(p) \leq P_{th}} \log (1 - P(p)) \right)$$

where $P(p)$ is the on-plane confidence prediction for the point $p$, $O(S)$ is the on-plane point sets for symmetry plane $S$, i.e., those points within distance $\epsilon_1$ of $S$ and $Sym$ are all ground-truth symmetry planes for an occluded point cloud.

After some steps, some points start to have higher on-plane confidence prediction, then we are letting those points learn normal vectors directions. Compared to on-plane prediction, we loosen up the distance from $\epsilon_1$ to $\epsilon_2$. The loss for learning normal vectors is $L1$ loss between the prediction and the ground truth,

$$L_{nv} = \sum_{S \in Sym} \left( - \sum_{p \in O'(S) \wedge P(p) > P_{th}} \mathbb{L}_1(N(p) - N_{gt}(S)) \right)$$

where $N(p)$ is the normal vector prediction for point $p$, $N_{gt}(S)$ is the ground-truth normal vector of plane $S$, $O'(S)$ are those points within distance $\epsilon_2$ of $S$ and $Sym$ are all ground-truth symmetry planes for an occluded point cloud.

So the training loss, in conclusion, is the following,

$$L = L_{op} + \lambda L_{nv}$$

and $P_{th} = 0.6, \lambda = 10$ in our training.

## 3.3. Testing

Now that we are done with the training, now we have to fuse all symmetry plane predictions into final predictions as we briefed at the beginning of the section. We devise a three-step RANSAC in our inference stage:

1. We pick ten random points out of the point cloud and settle on the one point $C$ with the proposal sweeping across the maximum number of points.
2. According to the proposal of point $C$, we sift through those points with similar proposals ($L1$ distance $dist_1$) and take them out to prepare a unanimous vote. We denote the set as $Vote(C)$.
3. Initialized with the proposal of point $C$, an iterative optimizer is employed to generate a **proposal** that covers the maximum number of points. Besides the proposal, we also provide a **confidence score** calculated from the percentage of points covered in $Vote(C)$. After getting the proposal and the confidence score, points with similar proposals ($L1$ distance $dist_2$, we usually set $dist_2 > dist_1$) in the point cloud are scrubbed.

The three-step RANSAC is executed until there are insufficient number of points to feed into. Specifically, $dist_1 = 0.6$ and $dist_2 = 1.3$.

Through the procedure, we are able to generate some symmetry predictions and confidence scores, indicating how sure the algorithm is about the correctness of these predictions.

## 4. Experiments

In this section, we first conduct quantitative experiments comparing our performance on objects suffering from different degrees of occlusion with the state-of-the-art work from Ecins[5]. Next up, we show some of the galleries demonstrating prediction accuracies visually. All our experiments are conducted on the YCB-video dataset.

*4.1. Quantitative Accuracy Experiments*

We start by introducing our way of accuracy measurement. In our experiments, we use PR(precision-recall) curve to demonstrate the efficacy of our algorithm. Recall that precision = TP/(TP + FP) and recall = TP/(TP + FN). First off, let us expound the definition of true positives, false positives, and false negatives. We say a prediction $(c_1, \vec{n_1})$ is within the **correctness threshold** of a ground-truth symmetry plane $(c_2, \vec{n_2})$ if and only if two planes are close enough: $(c_2 - c_1)\vec{n_2} \leq d_{th}$ and have similar directions: $\angle(\vec{n_1}, \vec{n_2}) \leq angle_{th}$. The variable here for the PR curve is the **confidence threshold**, separating open predictions (those above) and suppressed predictions (those under).

**True positives(TP)** are those predictions within the correctness threshold and are confident enough (above the confidence threshold). **False positives(FP)** are predictions out of the correctness threshold but still have higher confidence than the confidence threshold. As for predictions within the correctness threshold but below the confidence threshold, they are regarded as **false negatives(FN)**. If there is a TP for a ground truth symmetry plane, then other predictions are ignored and are not considered as either FP or FN.

With the above definition in mind, we are going to break down accuracy performance according to two standards: intersection circumference and degree of occlusion. The degree of occlusion is the percentage of the occluded surface as opposed to the entire surface, including self-occlusion.

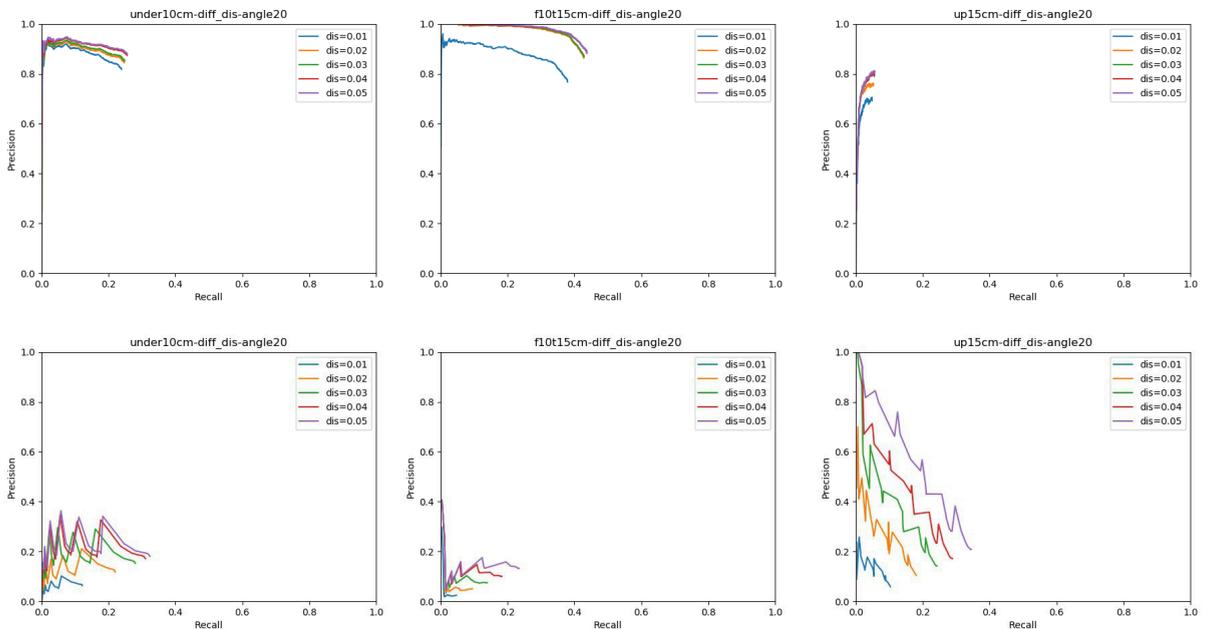

Fig. 2. PR curve comparison between ours(top) and Ecins[5](bottom) under different intersection circumferences

In Fig. 2, we show how our algorithm fares compared to Ecins[5] in different intersection circumferences. The thresholds are set as $d_{th} = 0.01, 0.02, ..., 0.05$ and $angle_{th} = 20°$. Intersection circumferences refer to the intersection between the symmetry plane and the object surface, i.e., the line on the surface that cut the object in half. Intuitively, we find for Ecins[5], which based on existing pixel correspondence, longer intersection circumferences mean we are likely able to find more correspondence, as shown in the bottom line in Fig. 4. That may explain why Ecins[5] has a sudden boost in performance in cases when intersection circumferences are longer than 15cm. As for cases where intersection circumferences are shorter than 15cm, the accuracy of Ecins[5] suffers considerably as opposed to our method.

The high accuracy of our algorithm could be led by the fact that in heavy occlusion, Ecins[5], which heavily relies on geometry information, is more susceptible to shape changes. Our algorithm, using deep learning, has acquired the capability to distinguish symmetry planes from the clutter.

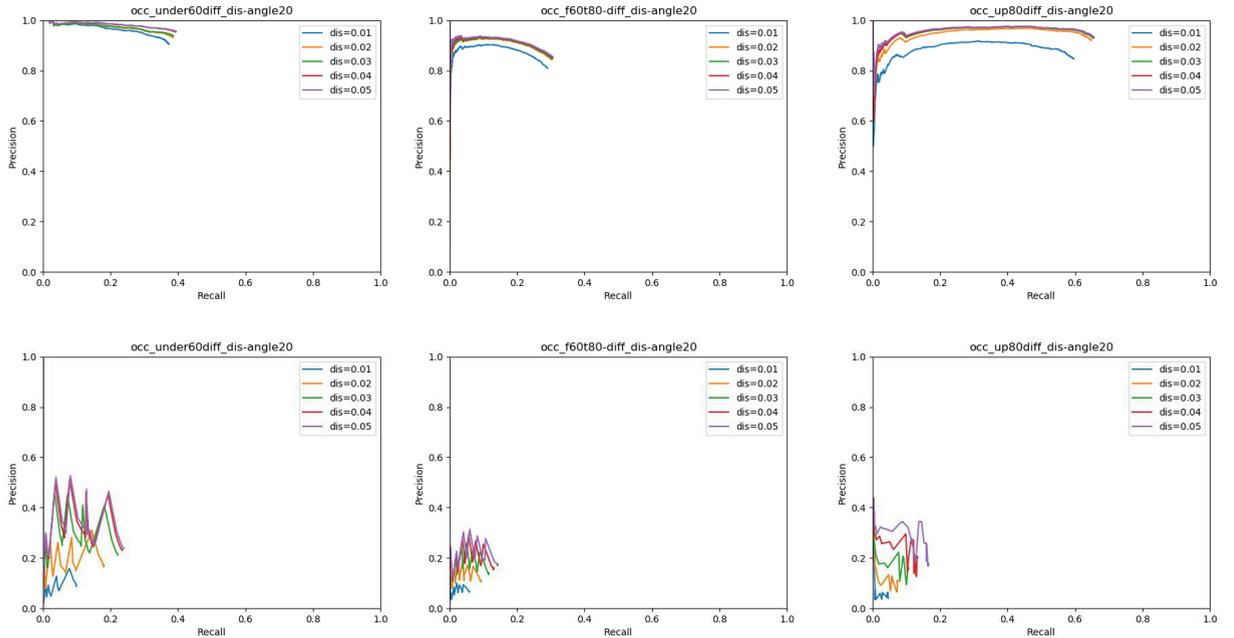

Fig. 3. PR curve comparison between ours(top) and Ecins[5](bottom) under different degrees of occlusion

Fig. 3 offers us another angle to compare our algorithm and Ecins. Under whatever degrees of occlusion, our algorithm surpasses Ecins in both recall rate and precision. Especially when it comes to large occlusions, our algorithm shows persistent accuracy and robustness in detecting symmetries from tiny unoccluded pieces. Thus, it is proper to say that deep learning could serve as a great aid in detecting symmetries in occluded scenarios.

Table 1. Precision boost when $dist_{th} = 0.01$. (Values are $\arg\max (prec \times recall)$)

| Degree of Occ. | Precision(Ecins) | Precision(Ours) | Recall(Ecins) | Recall(Ours) | P × R(Ecins) | P × R(Ours) |
|---|---|---|---|---|---|---|
| Under 60% | 0.16 | 0.91 | 0.08 | 0.38 | 0.013 | 0.346 |
| 60% to 80% | 0.09 | 0.80 | 0.04 | 0.32 | 0.004 | 0.256 |
| Above 80% | 0.06 | 0.86 | 0.04 | 0.60 | 0.002 | 0.516 |

In Table 1, we see how our algorithm performs under $dist_{th} = 0.01$ and $angle_{th} = 20°$. It concretely demonstrates that our algorithm has gained robustness towards reflective symmetry detection in occlusion in all categories.

### 4.2. Prediction Visualization

Now let us see how well our algorithm has dealt with each degree of occlusion split: under 60%, from 60% to 80% and above 80%.

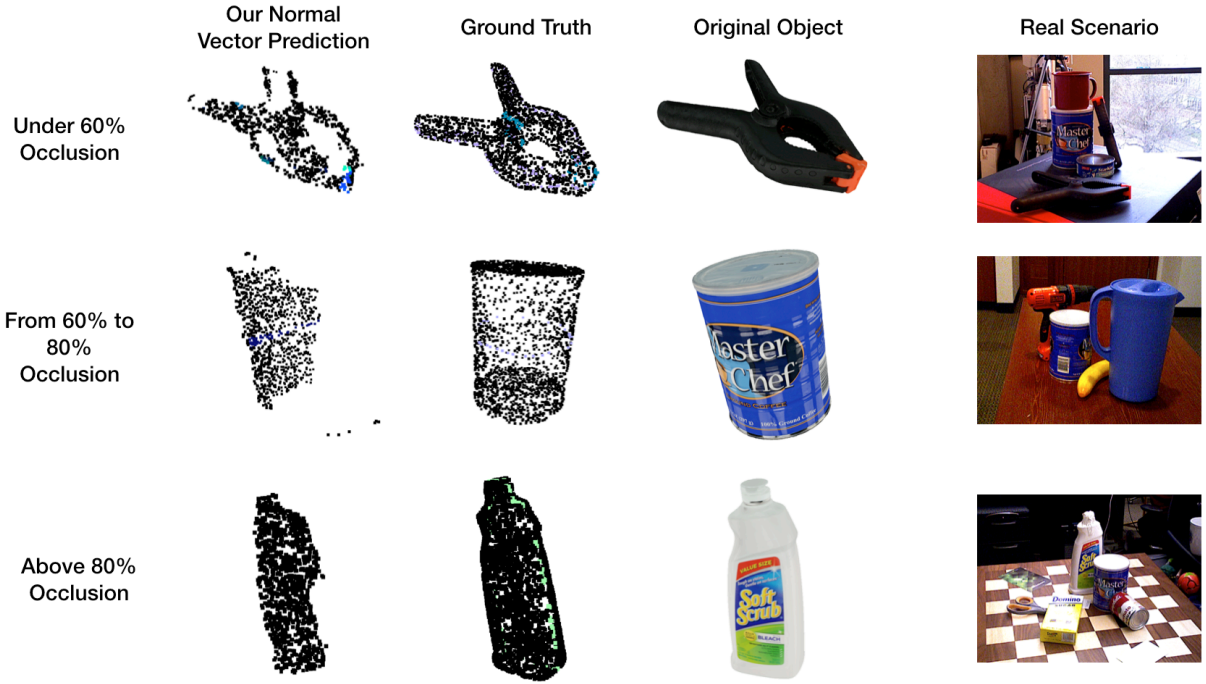

Fig. 4. Visualization of our predictions under different degrees of occlusion (Better viewed in PDF and zoomed in)

To get a clearer view of what happens in prediction, we need to take a look at Fig. 4. As we explained above, an RGB-D image is first segmented from a YCB-video frame and projected into 3D space as occluded point clouds. Through our later deep-learning framework, we are able to predict normal vectors for those on-plane points. Those highlighted points on the leftmost column are normal vector predictions by our algorithm. We have seen highlighted points align with ground truth planes, and the color of those points, normal vectors indicating different directions, is following unanimous predictions in each direction.

It is worth noted that for those symmetry planes with longer intersection circumferences like the bottom row, it is more likely the points on the symmetry plane are missing out, causing lower accuracy, as we demonstrated in Figure 2.

## 5. Conclusion

In this paper, we proposed a deep-learning framework to settle symmetry detection in a 3D scenario, specifically on point clouds. Objects may be cluttered, some heavily occluded, which are quite tricky for correspondence-based algorithms like Ecins5. Our framework has been proven by far superior to Ecins. When we divide the YCB-video dataset by degrees of occlusion, we found our framework excels both in recall rate and precision rate, meaning we can retrieve more symmetry planes and more accurately.

As far as we know, we are the first to leverage deep-learning techniques to tackle reflective symmetry on occluded 3D objects. Although our algorithm does require that at least part of the symmetry plane not occluded, it could inspire fellow researchers to drive toward a path where more robust and efficient deep-learning algorithms lie.

8